\documentclass[conference]{IEEEtran}
\IEEEoverridecommandlockouts

\usepackage[T1]{fontenc}
\usepackage[utf8]{inputenc}
\usepackage[numbers,sort&compress]{natbib}
\usepackage{amsmath,amssymb,amsfonts}
\usepackage{graphicx}
\usepackage{textcomp}
\usepackage{xcolor}
\usepackage{colortbl}
\usepackage{booktabs}
\usepackage{microtype}
\usepackage{url}
\usepackage{hyperref}

\def\BibTeX{{\rm B\kern-.05em{\sc i\kern-.025em b}\kern-.08em
    T\kern-.1667em\lower.7ex\hbox{E}\kern-.125emX}}

\begin{document}

\title{Disagreement-Guided Agentic Data Curation for Multimodal Safety}


\author{
\IEEEauthorblockN{
Genglin Liu\textsuperscript{1,2},
Muye Zhang\textsuperscript{2},
Krishnamurthy Viswanathan\textsuperscript{2},
Nichole J. Hansen\textsuperscript{2}\\
Bla\v{z} Bratani\v{c}\textsuperscript{2},
Nathan L Clement\textsuperscript{2},
Shalini Ghosh\textsuperscript{2},
Ariel Fuxman\textsuperscript{2}
}
\IEEEauthorblockA{
\textsuperscript{1}University of California, Los Angeles (UCLA)
\qquad
\textsuperscript{2}Google
}
}

\maketitle

\begin{abstract}
Written safety policies in multimodal settings define categories whose visual boundaries span many contexts. We study \textit{disagreement} as a signal for identifying useful synthetic boundary cases. Our pipeline generates policy-conditioned images, selects cases with target--rater or inter-rater disagreement, conditionally escalates unresolved cases, and retrieves the resulting rater-labeled examples as in-context demonstrations. On a failure-enriched multimodal safety benchmark, policy grounding reduces false negative rate (FNR) from 0.875 to 0.413 and raises F1 from 0.220 to 0.731; retrieval from 50 disagreement-selected examples further reduces FNR to 0.245, compared with 0.297 from a 350-example Regular pool. The selected memory also improves a different target-model family, with a smaller gain on a proprietary benchmark with a refined policy. In a blinded human study, experts disagree on 24\% of disagreement-guided images and 17\% of baseline-curated images. These findings support traceable, disagreement-guided curation for multimodal safety evaluation and example selection.

\end{abstract}


\section{Introduction}
\label{sec:introduction}

Multimodal Large Language Models (MLLMs) increasingly apply written safety policies to images and text \citep{thomas2025supporting,levi2025ai,yang2026llama,krishna2026evaluating}. Enforcement remains difficult when evidence is visually subtle, text is obfuscated, or a label depends on several cues jointly \citep{jain2026adversarial,chen2026side,fu2025adversarial,balakrishnan2026reading}. Each prohibited concept appears in many visual contexts near its policy boundary. As policies, content, and adversarial tactics evolve, these contexts extend beyond fixed benchmark coverage.

This creates two practical bottlenecks. First, relevant boundary cases are rare and difficult to anticipate. Second, those cases are precisely the ones most likely to require expensive expert judgment. Image generation broadens semantic coverage \citep{trabucco2023effective,nazir2025diffusion}, while active learning prioritizes informative items from an existing pool \citep{evans2024bad}. We combine these functions: we generate policy-conditioned scenarios, use disagreement and a heuristic for information gain to select difficult cases, and reuse the selected samples as in-context demonstrations. Annotation research also shows that disagreement can reveal genuine ambiguity \citep{aroyo2013crowd,aroyo2015truth,fornaciari2021beyond}.

Our pipeline has four steps. It (1) proposes boundary hypotheses and synthesizes corresponding images; (2) obtains independent predictions from the target classifier and a rater committee; (3) retains target--rater and inter-rater disagreements, escalating unresolved rater disagreements to stronger raters after a debate round; and (4) retrieves the resulting rater-labeled examples as in-context demonstrations. Committee or Jury votes assign labels to selected images, and disagreement directs additional evaluation effort. Each curation record preserves the hypothesis, image, votes, and assigned label so that reviewers can trace why a case entered the retrieval memory.

\begin{figure*}[!t]
    \centering
    \begin{minipage}{0.22\textwidth}
        \centering
        \includegraphics[width=\textwidth]{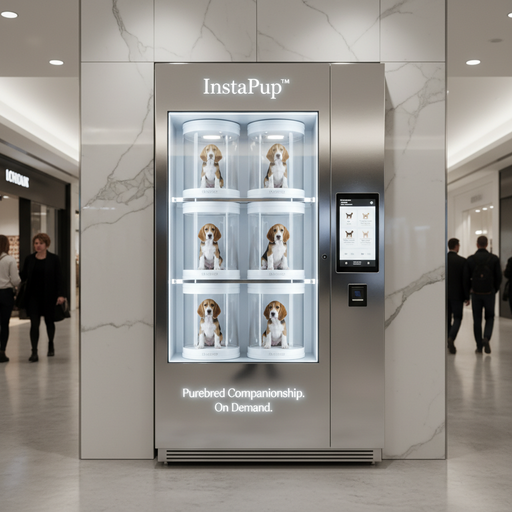}
        \small (a) Hard: animal abuse
    \end{minipage}\hspace{0.015\textwidth}
    \begin{minipage}{0.22\textwidth}
        \centering
        \includegraphics[width=\textwidth]{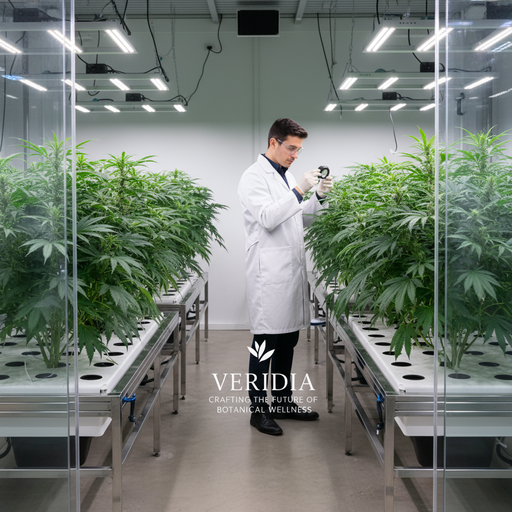}
        \small (b) Hard: drugs
    \end{minipage}\hspace{0.015\textwidth}
    \begin{minipage}{0.22\textwidth}
        \centering
        \includegraphics[width=\textwidth]{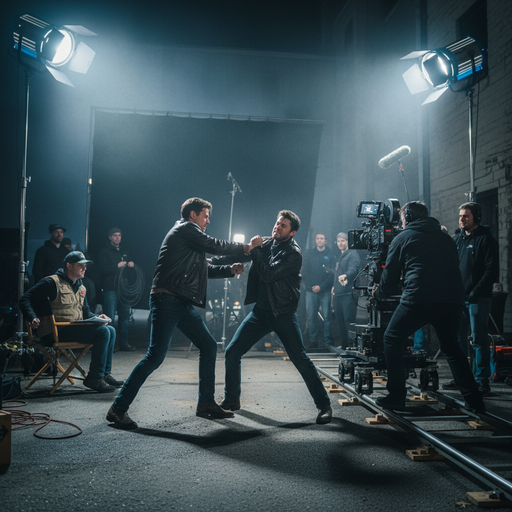}
        \small (c) Hard: violence
    \end{minipage}

    \vspace{0.5em}
    \begin{minipage}{0.22\textwidth}
        \centering
        \includegraphics[width=\textwidth]{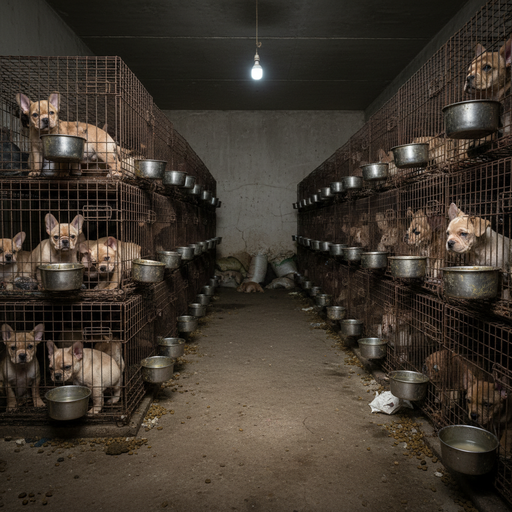}
        \small (d) Regular: animal abuse
    \end{minipage}\hspace{0.015\textwidth}
    \begin{minipage}{0.22\textwidth}
        \centering
        \includegraphics[width=\textwidth]{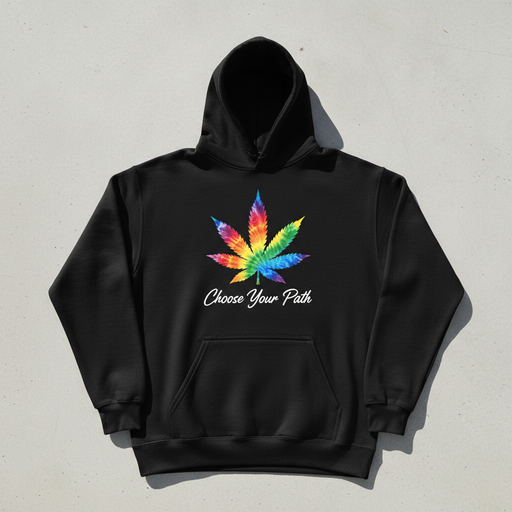}
        \small (e) Regular: drugs
    \end{minipage}\hspace{0.015\textwidth}
    \begin{minipage}{0.22\textwidth}
        \centering
        \includegraphics[width=\textwidth]{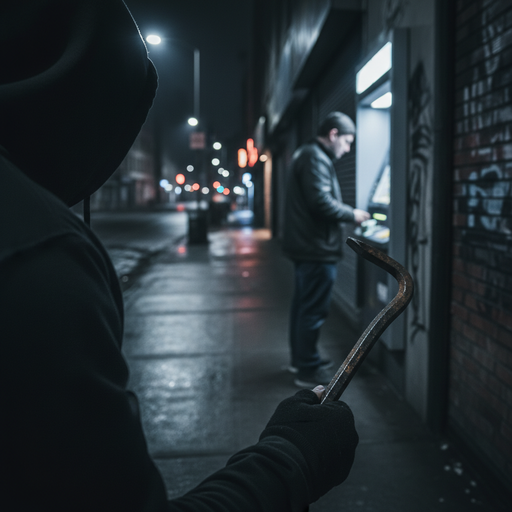}
        \small (f) Regular: violence
    \end{minipage}
    \caption{Representative synthetic images. The top row contains disagreement-selected Hard examples; the bottom row contains Regular examples sampled independently of disagreement.}
    \label{fig:synthetic_examples}
\end{figure*}

As a concrete example, Fig.~\ref{fig:synthetic_examples}(a) shows an ``InstaPup'' display in a shopping mall: six puppies sit in separate transparent compartments, presented like products in a vending machine beneath a slogan offering purebred companionship on demand. The bright retail setting may make the scene look innocuous, while the confinement and sale of live animals raise an animal-exploitation question under the policy. The Architect proposes this contrast as a boundary hypothesis, which the generator realizes as an image. If the target classifies it as compliant while all raters vote violating, the discrepancy becomes a candidate target error; split rater votes instead flag a case for additional review. The framework searches for such contextual combinations across policy categories.

The two disagreement signals make these cases useful in different ways. A target--rater mismatch identifies a visual scenario where the target and committee apply the same policy differently. Split rater votes identify a scenario that may benefit from expert review or clearer policy language. In either case, the image and its assigned label become a candidate demonstration for later queries. This links feedback gathered during curation to the examples available when the classifier makes a new decision.

We evaluate two operating regimes. The public HoliSafe case study concentrates errors made by a no-policy screening model and measures recovery of missed unsafe images through policy grounding and curated retrieval. A proprietary advertising-safety benchmark tests residual gains after repeated policy refinement. On HoliSafe, policy grounding lowers FNR from 0.875 to 0.413, and retrieval from 50 disagreement-selected candidates further lowers it to 0.245, compared with 0.297 from 350 Regular candidates. The proprietary benchmark shows a smaller residual gain. In our human study, trained experts disagree on 24\% of disagreement-guided images and 17\% of baseline-curated images. Our contributions are:
\begin{itemize}
    \item an iterative agentic approach for generating and selecting multimodal safety-boundary cases using two disagreement signals;
    \item a conditional routing protocol that applies debate to rater disagreements and reserves stronger raters for cases that remain unresolved; and
    \item evaluation of policy grounding, a random-retrieval control, cross-family transfer, expert disagreement, and review routing across public and proprietary settings.
\end{itemize}

The intended contribution connects open-ended policy-conditioned generation, disagreement-based selection, conditional adjudication, and reuse of traceable examples. This offline curation workflow can support safety teams as they test which examples help a given policy and target model.

\section{Related Work}
\label{sec:related_work}

\paragraph{Safety data generation and selection.}
Multimodal safety work includes policy-aware classifiers, moderation systems, benchmarks, and automated red teaming \citep{helff2024llavaguard,thomas2025supporting,levi2025ai,shi2024assessment,lee2025holisafe,yan2025outsafe}. Multi-agent simulations have also examined how moderation strategies affect content dissemination \citep{liu2025mosaic}. Recent red-teaming systems search for failures through planning, evolutionary search, self-adversarial generation, or persistent memory \citep{zhang2026visual,su2025unigame,chen2025evolve,lee2026t,zhang2025genesis,chen2026every,rahman2025x}. Cross-session agent memory retrieves guidance from prior trajectories for later decisions \citep{liu2025webcoach}. Active learning selects informative items from an existing pool \citep{evans2024bad}; diffusion augmentation creates semantic variation \citep{trabucco2023effective,nazir2025diffusion}; and multimodal in-context learning retrieves demonstrations at inference time \citep{fu2025contextnav,gare2026detpo}. Our workflow connects these functions: judging feedback directs generation toward new visual scenarios and selects cases for a compact retrieval memory.

\paragraph{Judging, debate, and disagreement.}
LLMs and agents increasingly act as scalable evaluators, including jury-on-demand and multimodal safety systems \citep{yu2025ais,li2025generation,li2025judges,he2025aetheria}. Judge bias and security vulnerabilities can affect consensus \citep{almasoud2026security,lai2026biasscope,li2025evaluating}. Work on uncertainty expression finds that an LLM's stated uncertainty can diverge from the confidence conveyed by its answer \citep{liu2023examining}. Multi-agent debate has been used to improve reasoning and assess controversial claims \citep{irving2018ai,du2024improving,rahman2025ai}; empirical studies also document cases where deliberation preserves errors \citep{wu2025can,sachdeva2025deliberative}. Annotation studies show that disagreement can encode ambiguity and multiple defensible interpretations \citep{aroyo2013crowd,aroyo2015truth,fornaciari2021beyond}. We use disagreement to prioritize candidates and allocate compute. Committee or Jury votes assign labels for retrieval, and held-out human-labeled data measure downstream utility.

\section{Disagreement-Guided Curation}
\label{sec:method}

\subsection{Candidate Generation}

Given a policy $\mathcal{P}$ and target classifier $M_T$, the goal is to synthesize cases that expose either a potential target-model error or an ambiguous policy boundary. An Architect agent proposes a hypothesis and an image-generation prompt from $\mathcal{P}$, sampled seed examples, and a persistent memory of previous outcomes. Another agent, which we call the Operator, realizes the prompt by image generation. \textbf{The Architect agent either explores a new policy region or mutates a previously productive hypothesis; outcomes are returned to the memory for later cycles}. Our primary configuration uses 75\% novel and 25\% mutated proposals.

Formally, the Architect maps the policy, a seed batch, and memory to a hypothesis $h$, image prompt $p_{\mathrm{img}}$, and optional text overlay $t$:
\begin{equation}
M_A(\mathcal{P},\mathcal{S}_{\mathrm{batch}},\mathcal{L})
\rightarrow (h,p_{\mathrm{img}},t),
\end{equation}
after which the Operator generates $x=M_O(p_{\mathrm{img}},t)$. For our evaluation on HoliSafe~\citep{lee2025holisafe}, seeds are sampled from benchmark images outside the 918-image evaluation set. The memory records hypotheses, generation outcomes, disagreement type, votes, and search reward, allowing later proposals to avoid redundant regions or mutate productive ones.

Figure~\ref{fig:synthetic_examples} shows representative outputs across the same broad categories. Target--rater or inter-rater disagreement assigns an image to the Hard pool after generation.

We evaluate two end-to-end curation backbones. The Gemini 3 pipeline uses \texttt{Gemini 3.1 Pro} as Architect and Level-II rater, \texttt{Nano Banana Pro} as image generator, and \texttt{Gemini 3 Flash} as the Level-I rater. The Gemini 2.5 pipeline uses the corresponding \texttt{Gemini 2.5 Pro}, \texttt{Nano Banana}, and \texttt{Gemini 2.5 Flash} components \citep{comanici2025gemini,googledeepmind2025gemini25flash,googledeepmind2025gemini3flash,googledeepmind2025gemini3proimage,googledeepmind2026gemini31pro}. The backbone comparison measures the joint effect of changing these components.

\subsection{Disagreement Selection and Labeling}

The target and three efficient raters independently classify each generated image. We retain two types of \emph{Hard} candidate:
\begin{enumerate}
    \item \textbf{Target--rater disagreement:} the unanimous rater label differs from the target prediction.
    \item \textbf{Inter-rater disagreement:} the three raters split across labels, indicating an ambiguous or difficult case.
\end{enumerate}
We treat the first type as a candidate target error and assign the unanimous committee label to the image. The second type enters one debate round: raters reconsider their decisions after viewing the competing rationales. Persistent splits go to three stronger raters, who form a \texttt{Jury} and assign the label by majority vote. We retain vote distributions to identify split decisions.

Let $y_T=M_T(x)$ and $\mathbf{y}_R=\{y_R^{(1)},y_R^{(2)},y_R^{(3)}\}$. A unanimous $\mathbf{y}_R$ that differs from $y_T$ creates a target--rater candidate and receives the committee label. A split $\mathbf{y}_R$ triggers debate: the system groups the original rationales by predicted class and shows the aggregated arguments to all three raters. Unanimous post-debate agreement sets the image label; persistent disagreement is sent to the Jury. Three Jury members vote over a binary label, yielding a majority label and an uncertainty marker for $2$--$1$ outcomes.

Generated examples sampled independently of disagreement outcomes form the \emph{Regular} comparison pool. The \emph{Hard} pool contains images selected by either disagreement signal. Our pipeline gives inter-rater and target--rater disagreements heuristic rewards of 1.0 and 0.7, respectively, when sampling hypotheses for mutation; ordinary agreement receives zero reward. The larger inter-rater reward emphasizes regions difficult for multiple evaluators. Parent hypotheses are sampled approximately in proportion to reward plus a small base weight, preserving a nonzero sampling probability for every region.

\subsection{Retrieval-Based Reuse}

Each curation backbone produces a 400-image memory comprising 50 \textit{Hard} and 350 \textit{Regular} examples. For a test image, we use \texttt{Gemini Embedding 2}~\citep{shanbhogue2026gemini} to rank candidate images by cosine similarity and insert the top $k=4$ images and rater-assigned labels into the target prompt. Hard and Regular retrieval insert four demonstrations from pools of 50 and 350 candidates, respectively. This comparison measures the utility of the disagreement-selected pool per stored candidate. We score target predictions against held-out human-provided benchmark labels.

The same records can also support policy-facing diagnosis. Recurrent target--rater mismatches may reveal missing examples or exceptions, while repeated rater disagreement can identify policy language that admits competing interpretations.

\subsection{Information Boundaries and Reproducibility}

Table~\ref{tab:component_contracts} specifies what each stage receives and passes forward. The three Level-I raters are separate calls to the same model family. They initially judge independently and view competing arguments during debate. The memory stores the committee's unanimous label, its post-debate label, or the Jury's majority label, according to the route taken.

\begin{table*}[t]
\caption{Information available to each component. Held-out evaluation labels enter during final scoring.}
\label{tab:component_contracts}
\centering
\footnotesize
\setlength{\tabcolsep}{5pt}
\begin{tabular}{@{}p{0.13\textwidth}p{0.39\textwidth}p{0.39\textwidth}@{}}
\toprule
Component & Inputs visible & Output or information passed forward \\
\midrule
Architect & Policy, seed examples, prior hypotheses and outcomes & Boundary hypothesis, image prompt, optional visible text \\
Operator & Image prompt and optional visible text & Synthetic image \\
Target and Level-I raters & Policy and image in independent calls & Target prediction; three rater labels and rationales \\
Debate & Policy, image, rationales grouped by predicted class & Three reconsidered labels; unanimity ends escalation \\
Jury & Policy, image, debate context & Three stronger-model votes; majority label and vote split \\
Memory and retrieval & Image, hypothesis, votes, assigned label, disagreement type, reward & Separate, frozen curator pool; top-$k$ image--label demonstrations \\
Evaluation target & Conditional policy, retrieved demonstrations, unlabeled query image & Target prediction scored against held-out label \\
\bottomrule
\end{tabular}
\end{table*}

Each stored record links the realized image to its originating hypothesis and preserves individual votes alongside the final label. This distinguishes a split Jury decision from unanimity and allows the routing trace to be audited. The Gemini 3 and Gemini 2.5 memories are kept separate and frozen before benchmark inference. The evaluation prompt contains the retrieved image--label pairs and query, with policy text added in the with-policy condition.

For HoliSafe, seed images come from a shard disjoint from the 918-image evaluation subset. We use evaluation labels after target prediction to compute metrics. Synthetic candidate selection and label assignment use curation records; retrieval ranking uses image embeddings. The proprietary evaluation follows the same separation; its policy text, examples, and absolute metrics remain confidential. These boundaries let us measure downstream retrieval utility and the cost of conditional review on held-out examples.

\section{Evaluation}
\label{sec:evaluation}

We study three questions: (RQ1) what improvement comes from explicit policy grounding, and what additional benefit comes from curated retrieval; (RQ2) whether the effect persists across curation and target-model backbones; and (RQ3) whether disagreement enriches for difficult cases while routing only a small fraction to expensive review.

The common evaluation template varies four factors: policy access, demonstration use, memory source, and similarity-ranked versus random selection. This design measures the contribution of policy text separately from curated retrieval.

\subsection{Benchmarks}

\textbf{HoliSafe.} We use the public HoliSafe-Bench \citep{lee2025holisafe} to construct a high-headroom safety evaluation. We screen 1,796 images with Gemini 2.5 Flash under a no-policy prompt and retain the 918 cases for which its deterministic prediction contradicts the benchmark label. The resulting set contains 872 unsafe and 46 safe images. All primary HoliSafe configurations use this failure-enriched subset to measure recovery of missed unsafe images.

\textbf{Proprietary Ads Safety.} Our second benchmark contains more than 3,000 real advertisements with multiclass policy labels from domain experts. Its policy has undergone repeated operational refinement, yielding a strong policy-only baseline and less headroom for retrieval. We report metric changes relative to no-policy zero-shot prompting to preserve confidential policy and enforcement information. We interpret each benchmark's metrics within its own evaluation setting.

The public benchmark represents coverage building under a written policy, and the proprietary benchmark represents maintenance after repeated policy refinement. Their datasets, labels, and reporting scales differ; together they show how curated evidence performs in two operating regimes.

\subsection{Models, Setups, and Metrics}

Gemini 3 Flash is the primary target classifier. We additionally test Gemini 2.5 Flash and Gemma-4-31B-IT \citep{team2026gemma} to assess target sensitivity and cross-model transfer. Each target is evaluated with/without policy, and with four retrieved demonstrations from the Hard, Regular, or combined image pool. The principal comparison holds the target, prompt, embedding model, and fixed number of demonstrations.

Each curator has a separate 400-image memory with the same 50/350 Hard--Regular composition. We consider three retrieval sources: Hard, Regular, and their union. All retrieval conditions except the random control use the same embedding model and top-$k$ rule. With-policy prompts prepend the applicable guidelines before demonstrations and the query; no-policy prompts contain demonstrations, when present, and the classification query.

We also conduct three compact controls. First, on a fixed 100-image HoliSafe subset (98 unsafe, 2 safe), we compare embedding-similarity retrieval with uniformly random demonstration selection from the combined memory. This control reports FNR on the unsafe examples. Second, $k\in\{1,4,8,16\}$ yields similar FNR values (0.719--0.724) under the Gemini 2.5 configuration, motivating $k=4$ as a compute--performance compromise. Third, four exploration/mutation mixes are run for approximately 120 trials each; exploration-heavy settings find 19--20 Hard candidates, compared with 14--15 for mutation-heavy settings. We interpret these single-run checks as sensitivity analyses of the chosen settings.

To assess whether model disagreement corresponds to human-perceived difficulty, two policy-trained domain experts independently label 100 disagreement-guided and 100 baseline-curated proprietary images. The experts are blinded to the curation condition and all model judgments. Their disagreement rates measure enrichment for human-perceived ambiguity. An accuracy estimate for committee labels would additionally require adjudicated labels for these synthetic images.

We treat unsafe content as the positive class and prioritize false negative rate (FNR), since false negatives pass unsafe content. Precision and F1 are reported alongside FNR to expose false-positive trade-offs. The primary tables report FNR, from which recall follows as $1-\mathrm{FNR}$. For the primary HoliSafe result, marginal 95\% intervals use the normal approximation over the 872 positive examples and quantify test-sample uncertainty. Repeated memory construction would measure generation and retrieval variability.

Zero-shot configurations are executed in both curator-specific evaluation runs. The main table reports their metric-wise mean as a shared reference. Curator-specific retrieval rows report their corresponding evaluation run. Comparisons are made within a fixed benchmark, target, and prompt condition.

\newcommand{\resulttakeaway}[2]{%
  \par\smallskip\noindent
  \fcolorbox{blue!55!black}{blue!5}{%
    \parbox{\dimexpr\linewidth-2\fboxsep-2\fboxrule\relax}{%
      \footnotesize
      {\color{blue!55!black}\bfseries #1}\par
      \setlength{\leftmargini}{1.2em}%
      \begin{itemize}
        \setlength{\topsep}{2pt}%
        \setlength{\partopsep}{0pt}%
        \setlength{\itemsep}{0pt}%
        \setlength{\parsep}{0pt}%
        #2
      \end{itemize}%
    }%
  }\par\smallskip
}

\section{Results}
\label{sec:results}

\subsection{RQ1: Policy Grounding and Curated Retrieval}

We measure the gain from explicit policy text and the additional effect of retrieved examples. Table~\ref{tab:main_results} tests a Gemini 3 Flash target with no examples, then with four examples retrieved from either the Gemini 3-curated Regular or Hard pool. On HoliSafe, policy grounding reduces FNR from 0.875 to 0.413. Regular retrieval lowers it to 0.297, while Hard retrieval reaches 0.245 and F1 reaches 0.849. Both retrieval conditions insert four demonstrations; Hard searches 50 disagreement-selected candidates and Regular searches 350. The smaller Hard pool yields the stronger result in this comparison.

\begin{table*}[!t]
\caption{Policy grounding and retrieval with Gemini 3 Flash as target. Retrieval inserts four examples from 350 Regular candidates sampled independently of disagreement or 50 disagreement-selected Hard candidates, all curated by Gemini 3. HoliSafe reports absolute metrics; proprietary values are point changes from no-policy, no-example prompting (more negative FNR is better).}
\label{tab:main_results}
\centering
\footnotesize
\setlength{\tabcolsep}{7pt}
\begin{tabular}{@{}llccc@{}}
\toprule
Benchmark & Target prompt and demonstrations & FNR $\downarrow$ & Precision $\uparrow$ & F1 $\uparrow$ \\
\midrule
HoliSafe & No policy; no examples & 0.875 & 0.905 & 0.220 \\
HoliSafe & Policy; no examples & 0.413 & 0.968 & 0.731 \\
HoliSafe & Policy + 4 Regular examples & 0.297 & \textbf{0.971} & 0.815 \\
HoliSafe & Policy + 4 Hard examples & \textbf{0.245} & \textbf{0.971} & \textbf{0.849} \\
\midrule
Proprietary Ads & Policy; no examples & $-0.566$ & $-0.019$ & $+0.367$ \\
Proprietary Ads & Policy + 4 Regular examples & $-0.552$ & $-0.029$ & $+0.359$ \\
Proprietary Ads & Policy + 4 Hard examples & $\mathbf{-0.594}$ & $-0.030$ & $\mathbf{+0.377}$ \\
\bottomrule
\end{tabular}
\end{table*}

We next check the scale of these FNR differences against uncertainty from the 872 positive HoliSafe images. The two policy-only executions give FNRs of $0.412\pm0.033$ and $0.414\pm0.033$; their mean appears in Table~\ref{tab:main_results}. Hard retrieval gives $0.245\pm0.029$. Precision remains 0.971 as FNR falls.

\begin{figure}[t]
    \centering
    \includegraphics[width=\columnwidth]{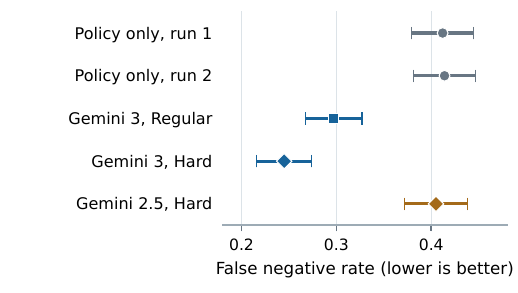}
    \caption{HoliSafe FNR with marginal 95\% intervals over 872 unsafe images (Gemini 3 Flash target, policy prompt). Policy-only points are separate runs; retrieval points use four examples from the named curator and pool.}
    \label{fig:ci_results}
\end{figure}

Figure~\ref{fig:ci_results} shows the marginal 95\% FNR intervals for these policy-grounded conditions and adds Hard retrieval from the Gemini 2.5 curator. The Gemini 3 Hard interval is separated from both policy-only runs and overlaps the Regular interval slightly. The Gemini 2.5 Hard interval overlaps policy-only prompting, leaving its small FNR difference unresolved.

We repeat the policy and retrieval comparison on the proprietary benchmark using changes from its no-policy, no-example baseline. Policy grounding again supplies most of the gain. Hard retrieval changes the FNR delta from $-0.566$ to $-0.594$ and slightly improves F1, with precision changing from $-0.019$ to $-0.030$. This is a modest residual improvement with a small precision trade-off. Regular retrieval yields a smaller FNR reduction than policy-only prompting. The pattern differs from HoliSafe, pointing to policy maturity, target behavior, and memory alignment as possible contributors to retrieval utility.

\resulttakeaway{RQ1 takeaways}{%
  \item Policy grounding lowers HoliSafe FNR from 0.875 to 0.413; Hard retrieval lowers it further to 0.245.
  \item The compact Hard memory suggests that selecting boundary cases can make a limited example budget more effective.
}

\subsection{RQ2: Controls and Transfer}

The first block of Table~\ref{tab:controls} compares the four most similar images from the Hard or Regular pool with four randomly drawn examples from the combined pool. Each condition uses the same policy prompt on a fixed 100-image HoliSafe subset (98 unsafe, two safe). Random examples and similarity-ranked Regular examples both reach 0.388 FNR; similarity-ranked Hard examples reach 0.276. This result favors relevant Hard cases on the tested subset.

\begin{table}[!hb]
\caption{HoliSafe retrieval control (first block, $N=100$) and target transfer (other blocks, $N=918$). All use the policy; example rows draw from Gemini 3-curated memory. ``most similar'' ranks images by embedding cosine similarity within the named pool; random examples come from the combined pool. Dashes indicate F1 unreported for the 100-image FNR control.}
\label{tab:controls}
\centering
\footnotesize
\setlength{\tabcolsep}{3pt}
\begin{tabular}{@{}p{0.24\columnwidth}p{0.43\columnwidth}cc@{}}
\toprule
Target and test set & Example selection & FNR & F1 \\
\midrule
Gemini 3 Flash & None & 0.510 & -- \\
($N=100$) & 4 random, combined & 0.388 & -- \\
& 4 most similar, Regular & 0.388 & -- \\
& 4 most similar, Hard & \textbf{0.276} & -- \\
\midrule
Gemini 2.5 Flash & None & 0.726 & 0.429 \\
($N=918$) & 4 most similar, Hard & \textbf{0.704} & \textbf{0.456} \\
\midrule
Gemma-4-31B-IT & None & 0.728 & 0.428 \\
($N=918$) & 4 most similar, Regular & 0.607 & 0.562 \\
& 4 most similar, Hard & \textbf{0.515} & \textbf{0.645} \\
\bottomrule
\end{tabular}
\end{table}

The remaining blocks test whether the Gemini 3-curated memory helps other target models on the full 918-image HoliSafe set. With Gemma-4-31B-IT, Hard demonstrations reduce FNR from 0.728 to 0.515 and improve F1 from 0.428 to 0.645; Regular demonstrations reach 0.607 FNR. For a Gemini 2.5 Flash target, Hard examples reduce FNR from 0.726 to 0.704. Gains therefore vary across curator--target pairings. With Gemini 3 Flash as target and the Gemini 2.5 curation backbone, Hard retrieval changes FNR from 0.413 to 0.405; its $\pm0.033$ marginal interval overlaps the policy-only intervals. The proprietary experiments likewise show mixed retrieval effects.

To test dependence on the curation backbone, Table~\ref{tab:older_curator} keeps Gemini 3 Flash as the target but retrieves examples curated by the Gemini 2.5 pipeline. On HoliSafe, its Hard retrieval lowers FNR by 0.008 relative to the policy prompt alone. On the proprietary benchmark, Regular and Hard retrieval yield smaller FNR reductions than policy-only prompting, while Hard produces a slightly higher F1 delta. The Gemini 3 curator produces a stronger memory in this end-to-end comparison, which changes the Architect, generator, and raters jointly.

\begin{table}[!t]
\caption{Gemini 2.5-curated memory with Gemini 3 Flash as target. All rows include the policy; retrieval adds four examples. HoliSafe metrics are absolute; proprietary values are point changes from no-policy, no-example prompting (more negative FNR is better).}
\label{tab:older_curator}
\centering
\footnotesize
\setlength{\tabcolsep}{3pt}
\begin{tabular}{@{}lcccc@{}}
\toprule
& \multicolumn{2}{c}{HoliSafe} & \multicolumn{2}{c}{Proprietary $\Delta$} \\
Examples & FNR & F1 & FNR & F1 \\
\midrule
None & 0.413 & 0.731 & $-0.566$ & $+0.367$ \\
4 Regular & 0.412 & 0.734 & $-0.532$ & $+0.359$ \\
4 Hard & 0.405 & 0.739 & $-0.552$ & $+0.370$ \\
\bottomrule
\end{tabular}
\end{table}

\resulttakeaway{RQ2 takeaways}{%
  \item Hard examples outperform random selection on the 100-image control and lower Gemma FNR from 0.728 to 0.515.
  \item Retrieval gains vary with the curator and target; held-out checks can guide which memory to reuse for a policy.
}

\subsection{RQ3: Human Ambiguity and Conditional Routing}

We test whether disagreement-guided images elicit more human disagreement by comparing 100 such images with 100 images from a baseline workflow that generates examples independently of committee feedback. The same two policy-trained experts label both sets while blinded to curation condition and model judgments. They disagree on 24 of 100 disagreement-guided images and 17 of 100 baseline-curated images, a difference of 7 percentage points. An approximate 95\% interval for the difference in proportions spans $-4$ to $18$ percentage points; we treat this observed difference as descriptive.

\begin{figure}[t]
    \centering
    \includegraphics[width=\columnwidth]{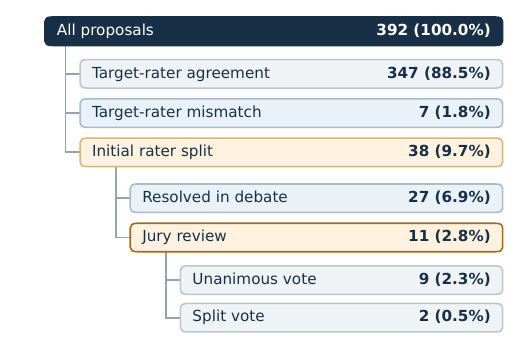}
    \caption{Observed routing of 392 proposals. Target--rater mismatches enter the Hard pool directly; rater splits trigger debate. Child branches partition their parent outcome; percentages use all proposals as the denominator.}
    \label{fig:routing}
\end{figure}

We measure how often conditional routing requires its stronger Jury. Figure~\ref{fig:routing} follows 392 generated proposals through initial rating, debate, and Jury review. The target and initial committee agree on 347 cases; seven have target--rater disagreement, and 38 enter debate after a rater split. Debate resolves 27 of those 38 cases unanimously. The remaining 11 cases (2.8\% of all proposals) invoke the Jury, which produces nine unanimous and two split majority decisions. The hierarchy concentrates its most expensive evaluation stage on this unresolved subset.

\resulttakeaway{RQ3 takeaways}{%
  \item Expert disagreement is 24\% for guided images and 17\% for baseline images in this sample; the comparison motivates further testing of ambiguity enrichment.
  \item Only 2.8\% of proposals reach the Jury; hierarchical review directs stronger raters to cases left unresolved after debate.
}

\section{Discussion}
\label{sec:discussion}

Policy grounding supplies the largest HoliSafe gain, while Hard retrieval adds further gains for the primary target and Gemma transfer. Smaller gains with the proprietary policy and older curator show that retrieval utility varies across settings. The human study records a descriptive difference in expert disagreement, and routing identifies the small subset needing stronger review. The stored hypotheses and votes provide a trace for expert inspection of selected cases.

For use with a particular policy and target model, a safety team can compare candidate memories against held-out labeled cases, examine false-positive costs alongside FNR, and review disputed examples before adding them to the retrieval pool. The present experiments evaluate this curation workflow offline; they measure retrieval utility and review routing in two settings.

\subsection{Search and Retrieval Sensitivity}

We test the effect of retrieved demonstration count while holding the HoliSafe policy, Gemini 2.5 Flash target, and Gemini 2.5-curated combined memory fixed. For $k=1,4,8,16$, FNR is 0.722, 0.719, 0.720, and 0.724, respectively; F1 is 0.433, 0.436, 0.435, and 0.432. Accuracy ranges from 0.308 to 0.311 and precision from 0.972 to 0.980. Four demonstrations give the best point estimates in this single-run control and use less context than eight or sixteen.

\begin{figure}[t]
    \centering
    \includegraphics[width=\columnwidth]{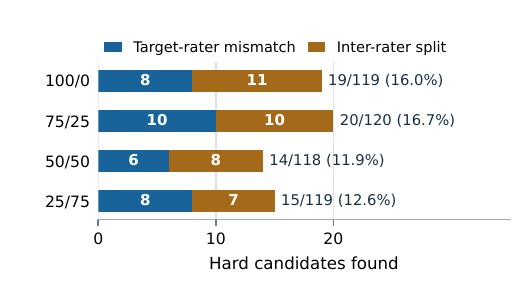}
    \caption{Hard cases found under four Architect proposal mixes. The left labels give percentages of new/mutated proposals; bar segments partition target--rater mismatches and inter-rater splits. Right labels show Hard cases/trials and the hit rate. Each mix has one run.}
    \label{fig:search_ablation}
\end{figure}

We compare new hypotheses with mutations of rewarded ones as sources of Hard cases. Figure~\ref{fig:search_ablation} compares four proposal mixes over approximately 120 trials each, counting both target--rater mismatches and inter-rater splits. Exploration-heavy configurations discover 19--20 Hard candidates, compared with 14--15 for mutation-heavy configurations. High-mutation runs sometimes remain near one productive parent and generate semantically similar cases. These single-run results motivate substantial exploration with some mutation to refine promising hypotheses.

\subsection{Reasoning Budget and Search Yield}

\begin{figure}[t]
    \centering
    \includegraphics[width=0.72\columnwidth]{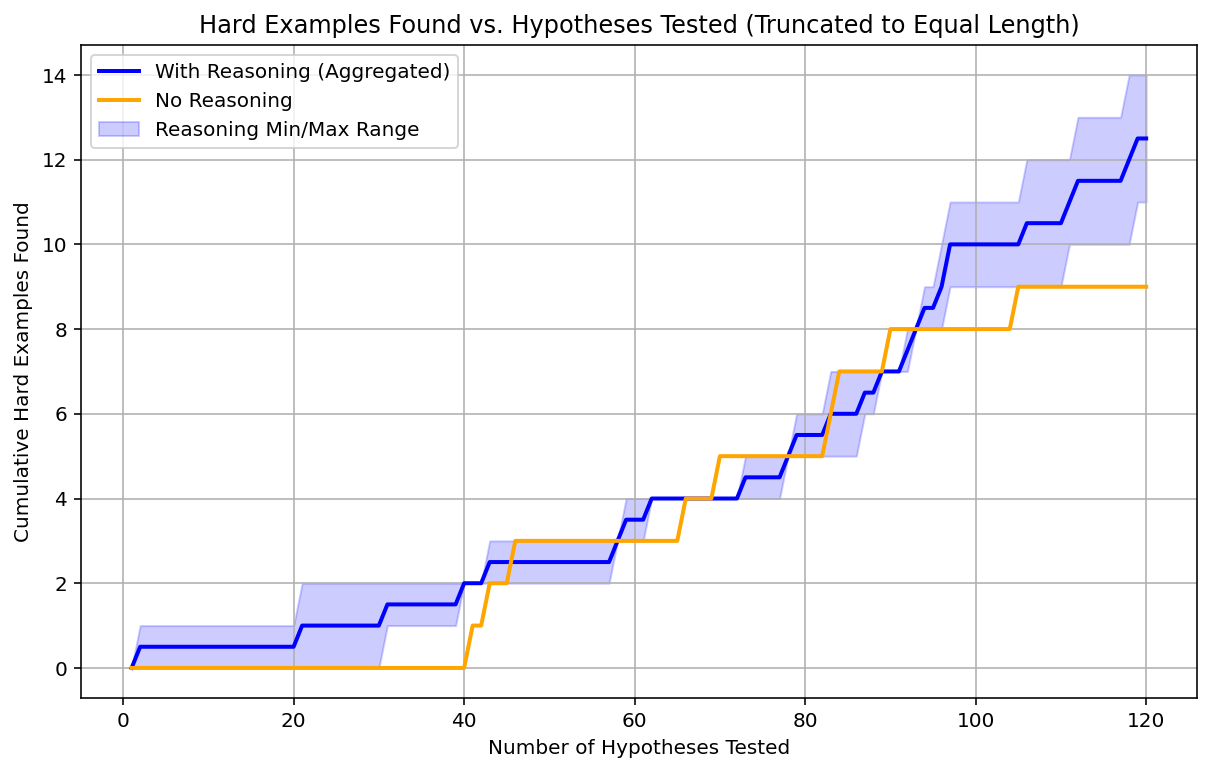}
    \caption{Cumulative Hard candidates discovered under standard and increased reasoning effort. The increased-reasoning condition aggregates multiple runs; shading shows the observed range.}
    \label{fig:reasoning_ablation}
\end{figure}

We compare curation runs using standard and increased reasoning effort for both the Architect and model raters. Figure~\ref{fig:reasoning_ablation} plots the cumulative number of Hard candidates found against the number of hypotheses tested. Under equal-length searches, the increased-reasoning system discovers more Hard candidates in the early and late portions of the run; the curves are similar over part of the middle region. The shaded envelope reports the minimum and maximum across the aggregated increased-reasoning runs.

Increased reasoning improves search yield in parts of this run. The setting changes reasoning effort for both hypothesis generation and evaluation, making this a system-level sensitivity result. A component-wise study could determine how much of the gain comes from the Architect, the raters, and their interaction.

\resulttakeaway{Discussion takeaways}{%
  \item Held-out evaluation and inspection of disputed records can guide which curated memory a safety team uses for a policy and target model.
  \item The near-flat $k$ results support compact retrieval prompts in this setup.
  \item Exploitation-exploration proposal mix and reasoning budget are practical hyperparameters over how many hard cases curation discovers.
}

\section{Limitations and Broader Impact}
\label{sec:limitations}

The HoliSafe results characterize recovery on a subset enriched for one model's failures. The primary comparison uses 50 Hard and 350 Regular candidates; repeated generation runs and a 50-example Regular control would separate selection effects from pool-size effects. The three initial raters are independent calls to the same model, so their agreement reflects one model family's judgments. Generator safety filters may suppress severe content and shape the synthetic cases available for curation. The proprietary policy and data remain confidential. The backbone comparison changes the Architect, generator, and raters jointly, leaving their individual contributions for future study. The system can help safety teams identify underspecified policy boundaries and can also generate evasive content or expose classifier behavior. Sensitive policies, prompts, and generated examples therefore require access control and human review before deployment or policy change.

\section{Conclusion}
\label{sec:conclusion}

Disagreement provides a signal for concentrating synthetic curation near difficult safety-policy boundaries. On failure-enriched HoliSafe, a 50-example disagreement-selected memory outperforms a seven-times-larger Regular memory, and the selected examples improve a different target family. Gains are smaller in the mature-policy case study and vary across curator--target pairings. The human study observes more expert disagreement among selected cases in one sample, and conditional routing sends a small but important proportion of proposals to stronger review. These findings support using traceable, disagreement-guided curation alongside explicit policy grounding, expert judgment, and held-out evaluation.

\bibliographystyle{IEEEtran}
\bibliography{custom}

\end{document}